\documentclass[11pt,a4paper]{article}

\usepackage{caption}
\usepackage{subcaption}
\usepackage{graphicx}
\graphicspath{{figures/}}

\usepackage{amssymb}
\usepackage{amsmath} % Required for some math elements 

\usepackage{algorithmic}
\usepackage{array}
\usepackage{lipsum}
\usepackage{hyperref}
\usepackage{natbib}
\usepackage{placeins}

\begin{document}

\title{A $\Delta$-evaluation function for column permutation problems} % Title
\author{J\'unior R. Lima, Vin\'icius Gandra M. Santos and Marco Antonio M. Carvalho$^*$\\Universidade Federal de Ouro Preto, Ouro Preto, MG, Brazil, 35400-000\\$^*$mamc@ufop.edu.br}

\date{\today} % Date for the report
\maketitle % Inserts the title, author and date

%----------------------------------------------------------------------------------------
%	Abstract
%----------------------------------------------------------------------------------------
\begin{abstract}
\normalsize
In this study, a new $\Delta$-evaluation method is introduced for solving a column permutation problem defined on a sparse binary matrix with the consecutive ones property. This problem models various $\mathcal{NP}$-hard problems in graph theory and industrial manufacturing contexts. The computational experiments compare the processing time of the $\Delta$-evaluation method with two other methods used in well-known local search procedures. The study considers a comprehensive set of instances of well-known problems, such as Gate Matrix Layout and Minimization of Open Stacks. The proposed evaluation method is generally competitive and particularly useful for large and dense instances. It can be easily integrated into local search and metaheuristic algorithms to improve solutions without significantly increasing processing time.

\end{abstract}

%----------------------------------------------------------------------------------------
%	Table of Content
%----------------------------------------------------------------------------------------
\setcounter{tocdepth}{2}
\tableofcontents

\clearpage

%----------------------------------------------------------------------------------------
%	Main Part
%----------------------------------------------------------------------------------------
\section{Introduction}
\label{intro}

The Column Permutation Problem (CPP), a complex and intriguing challenge in the realm of matrix operations and optimization problems, can be succinctly described as follows: given a sparse binary matrix, the task is to identify the corresponding permutation matrix that minimizes the maximum column sum, while adhering to the consecutive ones property.

Given a binary $m \times n$ matrix $P$,  $Q^\pi$ is the matrix obtained from a permutation $\pi$ of the $\{1, 2, \ldots, n\}$ columns of matrix $P$. $Q^\pi$ holds the \textit{consecutive ones property}, meaning all entries between two nonzero entries are also considered to have nonzero value in every row. The elements that switch values from zero to nonzero are called \textit{fill-ins}. A contiguous sequence of nonzero entries (fill-in or not) defines a \textit{1-block} in a row. Equation \eqref{eq1} defines the elements of matrix $Q^\pi$=\{$q^\pi_{ij}$\}, where function $\pi(i)$ returns the order of column $i$ in permutation $\pi$.

\begin{equation}
\label{eq1}
q^{\pi}_{ij} = 	\begin{cases}
				1, \mbox{ if } \exists x, \exists y | \pi(x) \leq j \leq \pi(y) \mbox{ and } p_{ix} = p_{iy} = 1\\
				0, \mbox{otherwise}.
			\end{cases}		
\end{equation}

\textit{Critical columns} have the largest sum in $Q^\pi$ (i.e., considering fill-ins) and represent the bottleneck of a solution to the CPP. Equation \eqref{eq2} defines the critical columns' value $Z_{CC}$($Q^\pi$).
 
 \begin{equation}
\label{eq2}
Z_{\mbox{CC}}(Q^\pi) = \max_{j \in \{1, \ldots, n\}}	\sum_ {i=1}^{m}	q^{\pi}_{ij}
\end{equation}

The CPP objective is to find a permutation $\pi \in \Pi$ of columns of $P$ such that the maximum critical column value is minimized in $Q^\pi$, vide Equation \eqref{eq3}.

 \begin{equation}
\label{eq3}
Z_{\mbox{CPP}}(P) = \min_{\pi \in \Pi}~Z_{\mbox{CC}}(Q^\pi)	
\end{equation}

Table \ref{tab1} presents an example of a $6 \times 6$ binary matrix $P$ (part $a$) and correspondent $Q^\pi$ for $\pi_1$=[5, 2, 4, 6, 3, 1] (part $b$) and $\pi_2$=[1, 6, 5, 4, 3, 2] (part $c$). Fill-ins are highlighted in boldface. $Z_{\mbox{CC}}(Q^{\pi_1}) = 3$ (columns 2, 4, 6 and 3 are critical) and $Z_{\mbox{CC}}(Q^{\pi_2}) = 6$ (columns 5 and 4 are critical). 

\begin{table}[!ht]
 \caption{$P$ and $Q^\pi$ for $\pi_1$=[5, 2, 4, 6, 3, 1] and $\pi_2$=[1, 6, 5, 4, 3, 2]. Fill-ins are highlighted in boldface.} \label{tab1}
 \centering
	\makebox[0pt][c]{\parbox{1\textwidth}{
	    \begin{minipage}[b]{0.33\hsize}\centering
		\begin{tabular}{ r|c | c | c | c | c | c |}
		\multicolumn{1}{r}{}  &  \multicolumn{1}{c}{1} & \multicolumn{1}{c}{2}& \multicolumn{1}{c}{3}& \multicolumn{1}{c}{4}& \multicolumn{1}{c}{5}& \multicolumn{1}{c}{6} \\
		\cline{2-7}
		1	&	1	&	1	&	0	&	0	&	0	&	0	\\ 
		\cline{2-7}
		2	&	1	&	0	&	1	&	0	&	0	&	0	\\ 
		\cline{2-7}
		3	&	0	&	0	&	0	&	1	&	1	&	0	\\ 
		\cline{2-7}
		4	&	0	&	0	&	0	&	1	&	0	&	1	\\ 
		\cline{2-7}
		5	&	0	&	1	&	0	&	0	&	1	&	0	\\ 
		\cline{2-7}
		6	&	0	&	0	&	1	&	0	&	0	&	1	\\ 
		\cline{2-7}
	\end{tabular}
        \caption*{($a$)}
    \end{minipage}
    \hfill
    \begin{minipage}[b]{0.33\hsize}\centering
	\begin{tabular}{r |c | c | c | c | c | c | }
		\multicolumn{1}{r}{}  &  \multicolumn{1}{c}{5} & \multicolumn{1}{c}{2}& \multicolumn{1}{c}{4}& \multicolumn{1}{c}{6}& \multicolumn{1}{c}{3}& \multicolumn{1}{c}{1} \\ 
		\cline{2-7}
		1	&	0	&	1	&	\textbf{1}	&	\textbf{1}	&	\textbf{1}	&	1	\\ \cline{2-7}
		2	&	0	&	0	&	0	&	0	&	1	&	1	\\ \cline{2-7}
		3	&	1	&	\textbf{1}	&	1	&	0	&	0	&	0	\\ \cline{2-7}
		4	&	0	&	0	&	1	&	1	&	0	&	0	\\ \cline{2-7}
		5	&	1	&	1	&	0	&	0	&	0	&	0	\\ \cline{2-7}
		6	&	0	&	0	&	0	&	1	&	1	&	0	\\ \cline{2-7}
		\cline{2-7}
	\end{tabular}
        \caption*{($b$)}
    \end{minipage}
    \hfill
    \begin{minipage}[b]{0.32\hsize}\centering	
	\begin{tabular}{r |c | c | c | c | c | c | }
 		\multicolumn{1}{r}{}  &  \multicolumn{1}{c}{1} & \multicolumn{1}{c}{6}& \multicolumn{1}{c}{5}& \multicolumn{1}{c}{4}& \multicolumn{1}{c}{3}& \multicolumn{1}{c}{2} \\ 
		\cline{2-7}
		1	&	1	&	\textbf{1}	&	\textbf{1}	&	\textbf{1}	&	\textbf{1}	&	1	\\ \cline{2-7}
		2	&	1	&	\textbf{1}	&	\textbf{1}	&	\textbf{1}	&	1	&	0	\\ \cline{2-7}
		3	&	0	&	0	&	1	&	1	&	0	&	0	\\ \cline{2-7}
		4	&	0	&	1	&	\textbf{1}	&	1	&	0	&	0	\\ \cline{2-7}
		5	&	0	&	0	&	1	&	\textbf{1}	&	\textbf{1}	&	1	\\ \cline{2-7}
		6	&	0	&	1	&	\textbf{1}	&	\textbf{1}	&	1	&	0	\\ \cline{2-7}
		\cline{2-7}
	\end{tabular}
        \caption*{($c$)}
    \end{minipage}
}}	
\end{table}

This generic problem models successfully a wide range of equivalent $\mathcal{NP}$-hard problems in a variety of contexts \citep{Mohring1990, Linhares2002}, including graph theory (Interval Thickness, Node Search Game, Edge Search Game, Narrowness, Split Bandwidth, Pathwidth, Edge Separation, Vertex Separation and Modified Cutwidth), very large scale integration design (Programmable Logic Array Folding and Gate Matrix Layout) and production planning (Minimization of Open Stacks). \cite{Linhares2002} present a rich discussion on the connections among these problems (and some others) and some theoretical results on the hardness of solving them.

Algorithms for permutation problems commonly employ combinations of insertion heuristics to explore different solutions, such as best insertion, 2-swap, and 2-opt. These heuristics consist of, given a partial or complete initial solution, inserting elements of the solution into different positions and choosing one according to a predefined criterion, usually based on the resulting solution value. Such heuristics are also used as local search methods in metaheuristics and require successive application of evaluation functions at each insertion move performed to reach a local optimum. Particularly in problems whose representation is made using matrices as the CPP, the many evaluation procedure calls become computationally very costly and may even become prohibitive, given the high processing time required for thoroughly evaluating an $m \times n$ matrix, which is bounded by $\Omega(mn)$. To reduce the asymptotic complexity, or at least the algorithm running time, evaluating only the portion of the solution that was changed in each move is desirable. These evaluation functions are called $\Delta$-evaluation or fast evaluation methods.

This study proposes a bitwise $\Delta$-evaluation for the CPP and presents a time comparison of this evaluation with two other evaluation methods. The evaluation methods are also applied to three well-known local search procedures (best insertion, 2-swap, and 2-opt) to compare the running times. To illustrate those local search procedures, two of the problems mentioned above were chosen: Gate Matrix Layout and Minimization of Open Stacks. The $\Delta$-evaluation is faster than the other two methods and has its main advantage in larger and more dense instances.

The remainder of this report is organized as follows. Section \ref{problems} gives an overview of the chosen problems.  In Section \ref{methods}, we introduce the evaluation methods. Section \ref{results} reports the computational experiments carried out, and conclusions are drawn in Section \ref{conclusions}.  

\section{Related column permutation problems}
\label{problems}

The next sections briefly describe each related problem considered in this study and give an overview of related studies. The problems are relevant to the industry and are still being studied today with recent meaningful research.

\subsection{Minimization of open stacks}
\label{mosp}
The problem of minimizing the maximum number of simultaneous open stacks \citep{Yuen1991} arises in industrial environments, in which a factory needs to supply specific combinations of products that have associated given demands. A single machine manufactures all the products in batches and handles a single product type at each stage. Whenever a customer orders a set of products, a new \textit{open stack} is associated with it, meaning that physical space around the machine is assigned to it until that order is fulfilled -- the \textit{stack's closure}. There is an implicit assumption of a physical limit on the space around the machine because there is not enough free room to simultaneously place all customers' orders. Thus, to better use that physical space, it is necessary to determine the sequence of the products' manufacturing.

The \textit{Minimization of Open Stacks Problem} (MOSP) is an $\mathcal{NP}$-Hard problem \citep{Linhares2002} defined on a binary sparse matrix $P$, where the $n$ rows correspond to the customers' orders and the $m$ columns correspond to each product type available. Entry $p_{ij} = 1$ if customer $i$ ordered product type $j$, otherwise, $p_{ij} = 0$. The consecutive ones property corresponds to the characteristic that a stack is considered open until it receives its last product, regardless of whether, at a given intermediary stage, it receives a product or not. The objective is to determine a permutation of products that minimizes the maximum number of simultaneous open stacks.

Constructive greedy heuristics for the MOSP were proposed by \cite{Yuen1991} and later improved in \cite{Yuen1995}. \cite{Faggioli1998} also proposed a greedy heuristic with an additional improvement phase performed by Tabu Search, which was later used along with Simulated Annealing by \cite{Fink1999}.  The \textit{Minimal Cost Node} heuristic, a greedy method based on a graph modeling proposed by \cite{Becceneri2004}, is still the current best-performing ad hoc heuristic up to date. A faster and more robust -- still less accurate -- heuristic based on graph search was proposed by \cite{Carvalho2014}. \cite{frinhani2018pagerank} proposed a fast PageRank-based heuristic for very large instances. The proposed heuristic performs better with large and less dense instances. The set of large instances is used in the experiments of this study as they are relevant for comparing the running times of the evaluation methods. 

Several different branch and bound procedures were proposed \citep{Yuen1995b, Yanasse1997b, Faggioli1998, Yanasse2004}, however, with no practical application due to running time restrictions. Other significant contributions include the identification of exceptional cases solvable in deterministic polynomial-time \citep{Yanasse1996}, a modeling using graphs \citep{Yanasse1997} and preprocessing operations \citep{Yanasse2010}. In 2005, the MOSP was chosen as the subject of an important constraint modeling challenge \citep{Challenge2005}, whose winner among many entries was a dynamic programming method proposed by \cite{Banda2007}. \cite{Chu2009} later improved this method by using a different search strategy and adding new pruning methods in a branch and bound method. \cite{martin2022mathematical} proposed two integer linear programming (ILP) formulations and a goal programming model for the MOSP. The ILP formulations were compared to three previous models from the literature, although none of the models clearly outperformed the others. Later, \cite{martin2023models} proposed an ILP for two- and three-stage stage two-dimensional cutting stock problems with a limited number of open stacks.

Concerning metaheuristics, the biased random-key genetic algorithm proposed in \cite{Resende2016} is the current state-of-the-art for the MOSP solution. The authors also identified that the MOSP solution space has a flat landscape and proposed a new evaluation function that allows the difference between two similar solutions. Lastly, a new local search procedure employed in a nested variable neighborhood descent was proposed in \cite{lima2017descent} and could match many optimal solutions.

\subsection{Gate matrix layout}

In the context of Very Large Scale Integration (VLSI) design, a \textit{Gate Matrix} is a bidimensional programmable logical device used to implement combinatorial logic circuits.  Different connections among OR and AND logical \textit{gates} are necessary to produce specific logical functions in those circuits. Each connection uses a \textit{wire} and a subset of gates called \textit{net}. Nets that do not share gates among them can be implemented into \textit{tracks} (i.e., physical rows). The circuit's underlying logic is not changed if the gates sequencing (i.e., their layout) is altered. The overall circuit area is a function of the number of tracks needed to implement the nets on the printed circuit. 

The \textit{Gate Matrix Layout Problem} (GMLP) is an $\mathcal{NP}$-Hard problem \citep{Kashiwabara1979} also defined on a binary sparse matrix $P$, where the $n$ rows correspond to the nets and the $m$ columns correspond to each gate. Entry $p_{ij} = 1$ if net $i$ connects gate $j$, and $p_{ij} = 0$ otherwise. The consecutive ones property corresponds to the characteristic that a wire spans from the first to the last gate it connects, meaning it might cross some other gates that do not belong to that specific net. The objective is to determine a permutation of the $m$ gates so that the number of tracks and the circuit area are minimized. For instance, the circuit represented in Table \ref{tab1} could be printed using three tracks containing net 1, nets 5, 4, and 2, respectively, and nets 3 and 6. According to \cite{Ferreira92}, the \textit{compaction step} (i.e., the grouping of nets in single tracks) can be performed in polynomial time.

Constructive heuristics have been designed for the GMLP since the late eighties. \cite{Wing1985} firstly modeled the problem using intersection graphs, \cite{Hwang1987} and \cite{Chen1988} used the \textit{min-net-cut} algorithm and \cite{Chen1990} and \cite{Hu1990} employed artificial intelligence techniques to outperform previous methods. In the next decade, the GMLP was predominantly addressed by metaheuristics and hybrid methods, such as Genetic Beam Search \citep{Shahookar1994}, Predatory Search \citep{Linhares1999a}, Microcanonical Optimization \citep{Linhares1999b}, Constructive Genetic Algorithms \citep{Oliveira2002} and a multi-population evolutionary approach using Memetic Algorithm and Genetic Algorithm as search engines \citep{Mendes2004}. 

Exact methods for the GMLP are found in the literature much less often. A dynamic programming formulation is provided in \cite{Deo1987}. However, experiments were not conducted. \cite{Giovanni2013} considered minimizing the number of tracks instead as a constraint to minimize the total wire length used and presented a branch and cut procedure. \cite{Resende2016}, which proposed a biased random-key genetic algorithm for the MOSP solution, also considered GMLP instances and reported good results. Currently, the state-of-the-art regarding GMLP is represented by an adaptive large neighborhood search with ad hoc heuristics proposed by \cite{santos2018adaptive}, whose study reports optimal results for a comprehensive set of GMLP and Minimization of Open Stacks instances.

\section{Methods}
\label{methods}

This section briefly describes the basic idea of the two basic evaluation methods. Furthermore, the proposed $\Delta$-evaluation method is presented in detail. The implementation in C++ is available at \href{https://github.com/MarcoCarvalhoUFOP/MOSPDeltaEvaluation.git}{GitHub} for noncommercial use.

\subsection{Complete matrix evaluation}
Matrices initially represent the problems presented in the previous section. To evaluate such problems, the most intuitive way is to go through the matrix every time a change in the solution is made. Hence, the evaluation of a solution using matrix representation is performed in $\Theta(nm)$ complexity. This operation is time-consuming and, depending on the size of the instance, it becomes impractical for, for example, the aforementioned local search procedures. In the worst-case scenario, the complexity of these procedures is bounded by $\Omega(n^3m)$ each when using the complete matrix evaluation. 

\subsection{Indirect evaluation}

The indirect evaluation method adopts a list to represent the input matrix. The idea is to store only the nonzero entries per column in a list $\lambda$ of $n$ elements, one representing each column. Each element of $\lambda$ is also a list, which stores the indices of rows containing nonzero entries in that column. Furthermore, this method relies on two support structures to evaluate the solution: an array that stores the number of nonzero entries per row and a Boolean array that, given a partial solution, informs which rows have 1-blocks being considered in the evaluation at each column being analyzed. In the worst-case scenario, the indirect evaluation method is bounded by $O(nm)$ complexity when the instance has no zero entries. Hence, this representation is particularly efficient when the input matrix is large and sparse. However, the complexity of the local search procedures is still bounded by $O(n^3m)$. 

\subsection{$\Delta$-evaluation}
Given an initial solution $\pi$, a \textit{move} removes a column from its \textit{original} position in $\pi$ and inserts it in a \textit{target} position. The insertion move, considering origin $<$ target, is carried by shifting the column forward one position at a time until the target position is reached. This shift can be interpreted as swapping a column on position $i$ with a column immediately to its right on position $i+1$. The behavior of the elements of $Q^\pi$ were grouped into four different sets:

\begin{itemize}
    \item \textbf{leading ones (L)}: nonzero entries at the beginning of a 1-block in a row;
    \item \textbf{trailing ones (T)}: nonzero entries at the end of a 1-block in a row;
    \item \textbf{consecutive ones (C)}: zero entries positioned in between nonzero entries in a row;
    \item \textbf{intermediate ones (I)}: nonzero entries in a row that neither are in the beginning nor the end of a 1-block. 
\end{itemize}

The representation of these sets was performed with the aid of a \textit{bitset} structure. This structure stores only one bit (value 0 or 1), indicating the presence or absence of elements in a particular set. 

Each column $i$ of the matrix $Q^\pi$ has its own sets $L_i$, $T_i$, $C_i$ and $I_i$. Consequently, there is a bit that represents for each stage if, for a given row, occurs or not leading or trailing ones, if there is a discontinuity in a 1-block, or if there is one that is neither a leading nor a trailing one. In this way, Boolean logic is applied bitwise in the sets of two columns affected by the swap move and is sufficient to evaluate the new solution value of $Q^\pi$ after the swap move.

It is crucial to emphasize that only the two stages related to the two columns being swapped will suffer changes on the previously mentioned sets. Hence, the value of the new solution after the swap is given without analyzing the entire matrix. The swap move and reevaluation of the solution will be carried out until the original column reaches the target position. When all the intermediary insertions are evaluated, it characterizes the \textit{best insertion} method in a given interval, which consists of analyzing the insertion of a given solution element in all possible positions of a solution.

From this point forward, the current column $i$ is the column that must be moved to the target position, and the next column $i+1$ is the column positioned immediately to the right of the current column. The following format represents the sets:

\begin{itemize}
    \item $L_{i}$ and $L_{i+1}$ -- Indicates the leading ones on stages $i$ and $i+1$;
    \item $I_{i}$ and $I_{i+1}$ -- Indicates the intermediate ones on stages $i$ and $i+1$;
    \item $T_{i}$ and $T_{i+1}$ -- Indicates the trailing ones on stages $i$ and $i+1$;
    \item $C_{i}$ and $C_{i+1}$ -- Indicates the consecutive ones on stages $i$ and $i+1$;
    
\end{itemize}

Adopting \textit{bitset} makes it possible to perform logical operations on all elements of a set with a single instruction. Thus, the row index is omitted for simplicity, considering only the column indices.

\subsubsection{Calculate $\Delta$}

The value of $\Delta$ indicates the difference between the original solution and the new solution obtained after changing the position of the columns. From the proposed approach, the $\Delta$ calculation is summarized in the following steps:

\begin{enumerate}
    \item Initialize the contents of each set considered for each column of $Q^\pi$. This procedure has complexity bounded by $O(nm)$ and is called only once;
    \item Simulate the insertion move in the solution and update the contents of each set considered for each column of $Q^\pi$ between the origin and the target position;
    \item Calculate the difference between the values of both solutions:
    \begin{itemize}
        \item If the result value is negative, the solution improved because the sum of the elements on the critical columns in the new solution is lower. At this point the solution is updated effecting the movement.
        \item Otherwise, the value of the critical columns increased or remained the same. In this case, the solution remains the same without performing the movement. 
    \end{itemize}
\end{enumerate}
 
All sets must be updated at each swap move of two consecutive columns. The update of each set, for the stages $i$ and $i+1$ are given by the following Boolean expressions: 

\begin{align}
L_i &=& &L_{i+1} \vee (L_i \wedge I_{i+1}) \vee (L_i \wedge T_{i+1})\\
L_{i+1} &=& &L_i \wedge (\neg I_{i+1} \wedge \neg T_{i+1})\\
I_i &=& &(T_{i+1} \wedge I_i) \vee (I_{i+1} \wedge \neg L_i)\\
I_{i+1} &=& &(L_i \wedge I_{i+1}) \vee (I_i \wedge \neg T_{i+1})\\
T_i &=& &T_{i+1} \wedge (\neg I_i \wedge \neg L_i)\\
T_{i+1} &=& &T_i \vee (T_{i+1} \wedge I_i) \vee (T_{i+1} \wedge L_i)\\
C_i &=& &(C_{i+1} \wedge \neg L_i) \vee T_i\\
C_{i+1} &=& &(C_i \wedge \neg T_{i+1}) \vee L_{i+1}
\end{align}

Note that the order in which those operations are processed is not important since each operation uses a copy of the original content of each set. Hence, if the $L_i$ set is updated first, the $L_i$ set used to calculate, for example, the set $C_i$ will be a copy of the original $L_i$ before its update. After calculating the new content of the eight sets, the original sets are updated.

\subsubsection{Update leading ones}
The leading ones set represents nonzero entries that start a 1-block in a particular stage. The stage in which the 1-block starts can directly influence the obtained result. Equation \eqref{eqOpen1} shows how to update the leading ones on stage $i$ when swapping it with stage $i+1$. In the sequence, each component of the equation is analyzed and explained. 

\begin{equation}
\label{eqOpen1}
L_i = L_{i+1} \vee (L_i \wedge I_{i+1}) \vee (L_i \wedge T_{i+1})
\end{equation}

\begin{itemize}
    \item $L_{i+1}$: The column referring to stage $i+1$ will become the column referring to stage $i$. Thus, leading ones on stage $i+1$ will become leading ones on stage $i$, starting the 1-block one column earlier;%, discarding any additional validation, since the stack had not been used in the $j$ previous stages (i.e., $j<i,\forall j \in \{1..i-1\}$);
    \item $(L_i \wedge I_{i+1})$: Express the relationship between leading ones on stage $i$ and the presence of intermediate ones on stage $i+1$. If both are true, the intermediate ones on stage $i+1$ will become the leading ones on stage $i$ and vice versa. If there are no leading ones on stage $i$, it is impossible to have intermediate ones on stage $i+1$. Thus, the swap will result in both zero entries. 
    \item $(L_i \wedge T_{i+1})$: Express the relationship between leading ones on stage $i$ and trailing ones on stage $i+1$. If a 1-block starts on stage $i$ and immediately ends on stage $i+1$, the swap of these two columns will result in no change in the leading ones on stage $i$. If there are leading ones on stage $i$ and no trailing ones on stage $i+1$, the leading ones will happen on stage $i+1$, and stage $i$ will be empty.
\end{itemize}

Based on Equation \eqref{eqOpen1}, it is possible to notice that the new leading ones on stage $i$ are a result of the relationship between leading ones, intermediate ones, and trailing ones on stage $i+1$ and the former leading ones on stage $i$.

Equation \eqref{eqOpen2} presents the Boolean expression to update the leading ones on stage $i+1$ when swapping it with stage $i$. The explanation of each component comes next.

\begin{equation}
\label{eqOpen2}
L_{i+1} = L_i \wedge (\neg I_{i+1} \wedge \neg T_{i+1})
\end{equation}

\begin{itemize}
    \item $(L_i \wedge \neg I_{i+1})$: Express the relationship between leading ones on stage $i$ and the absence of intermediate ones on stage $i+1$. The column referring to stage $i$ will be moved to stage $i+1$, thus, if there is no intermediate ones on stage $i+1$, the 1-block will start on stage $i+1$, instead of on stage $i$;
    \item $(L_i \wedge \neg T_{i+1})$: Certifies that the 1-block starting on stage $i$ is not immediately ending on stage $i+1$. If there are no trailing ones on stage $i+1$, the leading ones from stage $i$ will be moved to stage $i+1$.
\end{itemize}

Equation \eqref{eqOpen2} states that leading ones on stage $i+1$ are given by the leading ones on stage $i$ less the intermediate ones and trailing ones on stage $i+1$. 

\subsubsection{Update intermediate ones}
Intermediate ones represent rows that have already had their 1-block started in earlier stages but have not finished or ended. In other words, they are nonzero entries in a given column but not leading or trailing. Depending on the movements of columns, intermediate ones can fall at the beginning or end of a 1-block and, in turn, become a leading or trailing one.

The new intermediate ones on stage $i$ depend on the presence of trailing ones on stage $i+1$ and the absence of leading ones on stage $i$. Equation \eqref{eqInter1} demonstrates the necessary Boolean operations to update the intermediate ones on stage $i$ when swapping it with the column on stage $i+1$. 

\begin{equation}
\label{eqInter1}
I_i = (T_{i+1} \wedge I_i) \vee (I_{i+1} \wedge \neg L_i)
\end{equation}

\begin{itemize}
    \item $(T_{i+1} \wedge I_i)$: Demonstrates the relationship between trailing ones on stage $i+1$ and the presence of intermediate ones on stage $i$. Stage $i$ has intermediate ones, and stage $i+1$ has trailing ones. The two nonzero entries will only exchange places, and no change will be made to the presence of intermediate ones on stage $i$.
    \item $(I_{i+1} \wedge \neg L_i)$: Express the relationship between intermediate ones on stage $i+1$ and the absence of leading ones on stage $i$. Only intermediate ones on stage $i+1$ that do not have the 1-block starting on stage $i$ will continue to be intermediate ones on stage $i$ after the swap.
\end{itemize}

Equation \eqref{eqInter2} presents the Boolean expression to update the intermediate ones on stage $i+1$. The expression considers intermediate ones and trailing ones.  

\begin{equation}
\label{eqInter2}
I_{i+1} = (L_i \wedge I_{i+1}) \vee (I_i \wedge \neg T_{i+1})
\end{equation}

\begin{itemize}
    \item $(L_i \wedge I_{i+1})$: Demonstrate the relation between leading ones on stage $i$ and intermediate ones on the next stage. In the case of leading ones on stage $i$ and intermediate ones on stage $i+1$, the swap between these two stages will maintain the nonzero values on both stages. The leading ones on stage $i$ will become the intermediate ones on stage $i+1$ and vice versa.
    \item $(I_i \wedge \neg T_{i+1})$: Express the relation between intermediate ones on stage $i$ and absence of trailing ones on stage $i+1$. In this way, intermediate ones on stage $i$ will become intermediate ones on stage $i+1$ only if the 1-block is not ending on stage $i+1$.
\end{itemize}

In short, there will be an intermediate one on stage $i+1$ if there are leading ones on stage $i$ and intermediate ones on stage $i+1$ or if there are intermediate ones on stage $i$ and an absence of trailing ones on stage $i+1$.

\subsubsection{Update trailing ones}

Trailing ones represent nonzero entries that close a 1-block in a specific stage. After swapping stages, stages with trailing ones might turn into intermediate or even leading ones. Equation \eqref{eqClosed1} shows how to update trailing ones on stage $i$ when swapping it with stage $i+1$.

\begin{equation}
\label{eqClosed1}
T_i = T_{i+1} \wedge (\neg I_i \wedge \neg L_i)
\end{equation}

\begin{itemize}
    \item $T_{i+1}$: Initially, the new stage $i$ can only have trailing ones if there were trailing ones on the former stage $i+1$. Thus, the 1-block will end sooner. 
    \item $(\neg I_i \wedge \neg L_i)$: Determines the relationship between trailing ones on stage $i$ and the absence of intermediate ones and leading ones on stage $i$. In case there are no leading or intermediate ones on stage $i$, the trailing ones on stage $i+1$ are moved to stage $i$ after the swap.
\end{itemize}

The trailing ones on stage $i$ are defined by the trailing ones on stage $i+1$ and the intermediate ones and leading ones on stage $i$. If successful, this change results in a one-block ending one stage earlier.

To update the trailing ones on stage $i+1$, it is necessary to consider intermediate ones, trailing ones, and leading ones on stage $i$. Equation \eqref{eqClosed2} presents the Boolean expression to update the trailing ones on stage $i+1$.

\begin{equation}
\label{eqClosed2}
T_{i+1} = T_i \vee (T_{i+1} \wedge I_i) \vee (T_{i+1} \wedge L_i)
\end{equation}

\begin{itemize}
    \item $T_i$: 1-blocks that ended on stage $i$ will be postponed and end on stage $i+1$.
    \item $(T_{i+1} \wedge I_i)$: Express the relationship between trailing ones on stage $i+1$ and intermediate ones on stage $i$. 1-blocks that ended on stage $i+1$ and had intermediate ones on stage $i$, will continue to end on stage $i+1$ after the swap. The nonzero entries on both columns are swapped.
    \item $(T_{i+1} \wedge L_i)$: Determines the relationship between trailing ones on stage $i+1$ and leading ones on stage $i$. 1-blocks that started on stage $i$ and immediately ended on stage $i+1$ will not change after the swap. 
\end{itemize}

\subsubsection{Update consecutive ones}
The consecutive ones represent zero entries between two nonzero entries; these values are switched to nonzero entries and are also known as fill-ins. In other words, they are entries between leading ones and trailing ones that are not intermediate ones. Minimizing the number of consecutive ones directly affects the maximum value of critical columns per stage. Moving stages with consecutive ones often result in these consecutive ones disappearing. 

Given a swap of two stages, to update the consecutive ones on stage $i$, it is necessary to consider consecutive ones on stage $i+1$ and leading ones and trailing ones on stage $i$. Equation \eqref{eqConseq1} shows how to update the consecutive ones on stage $i$.

\begin{equation}
\label{eqConseq1}
C_i = (C_{i+1} \wedge \neg L_i) \vee T_i
\end{equation}

\begin{itemize}
    \item $(C_{i+1} \wedge \neg L_i)$: Determines the relationship between consecutive ones on stage $i+1$ and the absence of leading ones on stage $i$. After the swap of the two stages, consecutive ones on stage $i+1$ will continue to be consecutive ones on stage $i$ unless there are leading ones on stage $i$. In this case, the 1-block would start later, on stage $i+1$, and the consecutive ones on stage $i$ would cease to exist.
    \item $T_i$: For each trailing one on stage $i$, after the swap, the 1-block will have its ending postponed to stage $i+1$. This results in stage $i$ having a consecutive one. 
\end{itemize}

The analysis to update the consecutive ones on stage $i+1$ involves the consecutive ones on stage $i$ and the trailing and leading ones on stage $i+1$. Equation \eqref{eqConseq2} shows how to update the consecutive ones on stage $i+1$.

\begin{equation}
\label{eqConseq2}
C_{i+1} = (C_i \wedge \neg T_{i+1}) \vee L_{i+1}
\end{equation}

\begin{itemize}
    \item $(C_i \wedge \neg T_{i+1})$: Express the relationship between consecutive ones on stage $i$ and the absence of trailing ones on stage $i+1$. Consecutive ones on stage $i$ will also be carried to stage $i+1$ as consecutive ones unless the 1-block ends on stage $i+1$.
    \item $L_{i+1}$: Given an leading ones on stage $i+1$, the start of the 1-block on stage $i$ will result in a consecutive one on stage $i+1$, causing a zero entry between leading and trailing ones. 
\end{itemize}

\section{Computational results}
\label{results}

The computational experiments were conducted on an Intel Core i7-8700 3.2 GHz processor with 16 GB RAM using Ubuntu 18.04 LTS. The proposed methods were coded in C++ and compiled with GCC 7.4.0 and the -O3 optimization option. The implemetations are available for download at \href{https://github.com/MarcoCarvalhoUFOP/MOSPDeltaEvaluation.git}{GitHub} for noncommercial use.

Four different sets of artificial instances for MOSP and GMLP were considered. To better evaluate the processing time, the instance sets selected for the experiments are restricted to sets with larger instances. Firstly, \textit{Becceneri} instance set \citep{Becceneri2004} contains 710 instances ranging from 10 to 150 rows and columns. \cite{Chu2009} proposed an instance set, \textit{Chu\&Stuckey}, with 200 instances ranging from 30 to 125 rows and columns. The \textit{Larger\&Harder} instance set, proposed by \cite{Carvalho2014}, contains 150 instances with 150, 175, or 200 rows and columns. Finally, \cite{frinhani2018pagerank} proposed an even larger instance set, \textit{Very Hard}, containing 610 instances with 400, 600, 800, or 1000 rows and columns.

This study focuses on assessing each evaluation method's performance rather than achieving optimal solutions. Hence, three well-known and simple descent local search methods will be applied to a random initial solution. The procedures are the best insertion, first improvement 2-swap, and 2-opt. 

The procedures apply consecutive moves on the solution and preserve the move only in case of improvement of the solution value. Otherwise, the move is discarded. If the move improves the solution, the local search starts over again. The procedure stops when all possible moves have been considered, and no further improvement has been made. Disregarding restarts, the three presented procedures have neighborhoods, i.e., the number of possible moves, bounded by $O(n^2)$. Each instance was run ten independent times for each local search procedure. For each time, one numerically predetermined seed was used to generate random numbers; hence, for each instance, the three local searches found the same solution and number of improvements, leading to the procedure restart, allowing for a fair comparison. 

The time comparison presented in the following graphs for each local search procedure was converted into a logarithm scale for better data visualization. Moreover, the time values for instances bigger than 600$\times$600, in Figures (d), for the matrix evaluation method is an interpolation based on the size of the instance. Finally, for each local search procedure, statistical experiments were conducted between the $\Delta$-evaluation and the indirect evaluation method on the Very Hard instance set time results to analyze their different performance further. A standard test is performed first, Shapiro--Wilk normality test \citep{shapiro1965analysis}, and rejected the null hypothesis that the results could be modeled according to a normal distribution. Then, given that the results are not normally distributed, the non-parametric Wilcoxon signed-rank test \cite{rey2011wilcoxon} was employed to investigate if there is a significant difference between the time performance of the $\Delta$-evaluation and the indirect evaluation method.

\subsection{Best Insertion}
\begin{figure}[!ht]
\centering 
\minipage{0.45\textwidth}
  \includegraphics[width=\linewidth]{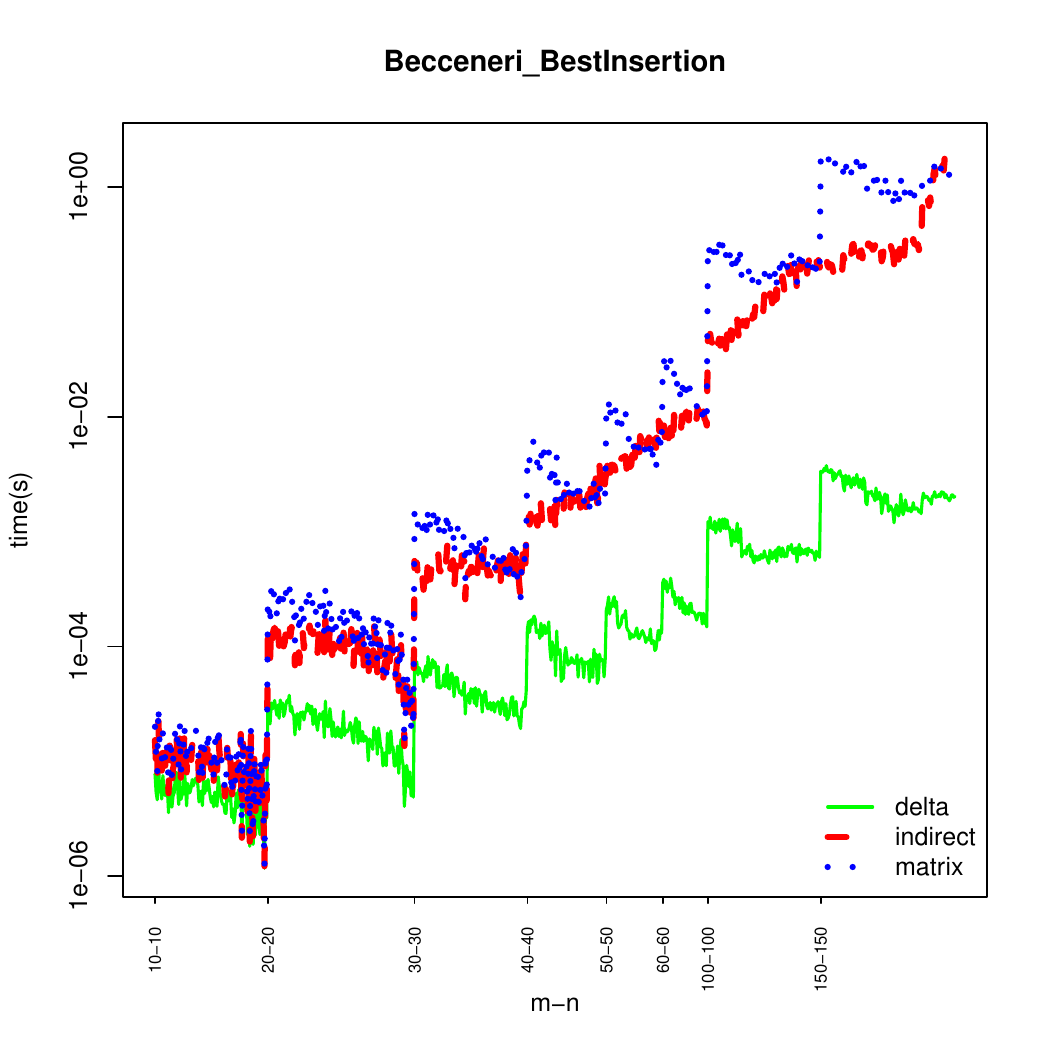}
  \caption*{(a)}
\endminipage
\minipage{0.45\textwidth}
  \includegraphics[width=\linewidth]{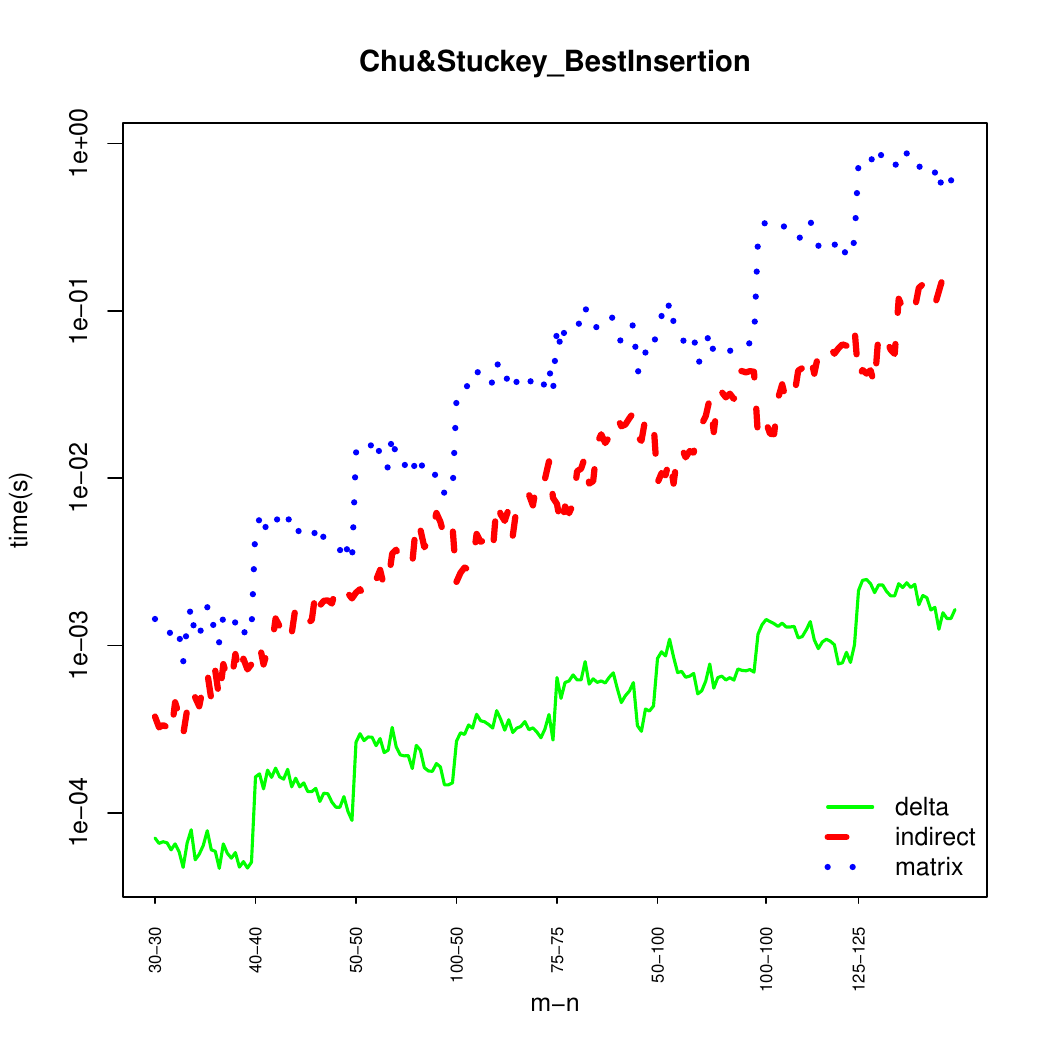}
  \caption*{(b)}
\endminipage
\\
\minipage{0.45\textwidth}
  \includegraphics[width=\linewidth]{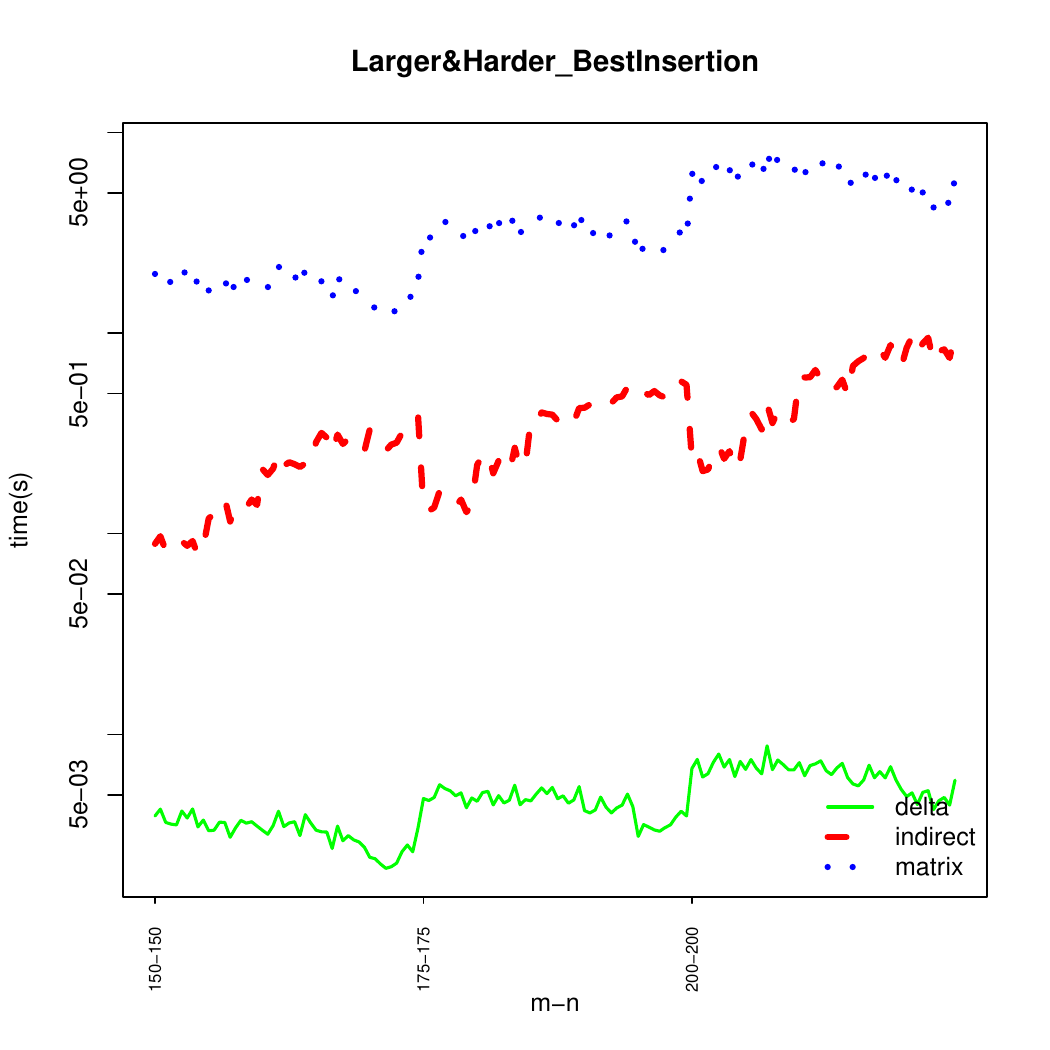}
  \caption*{(c)}
\endminipage
\minipage{0.45\textwidth}
  \includegraphics[width=\linewidth]{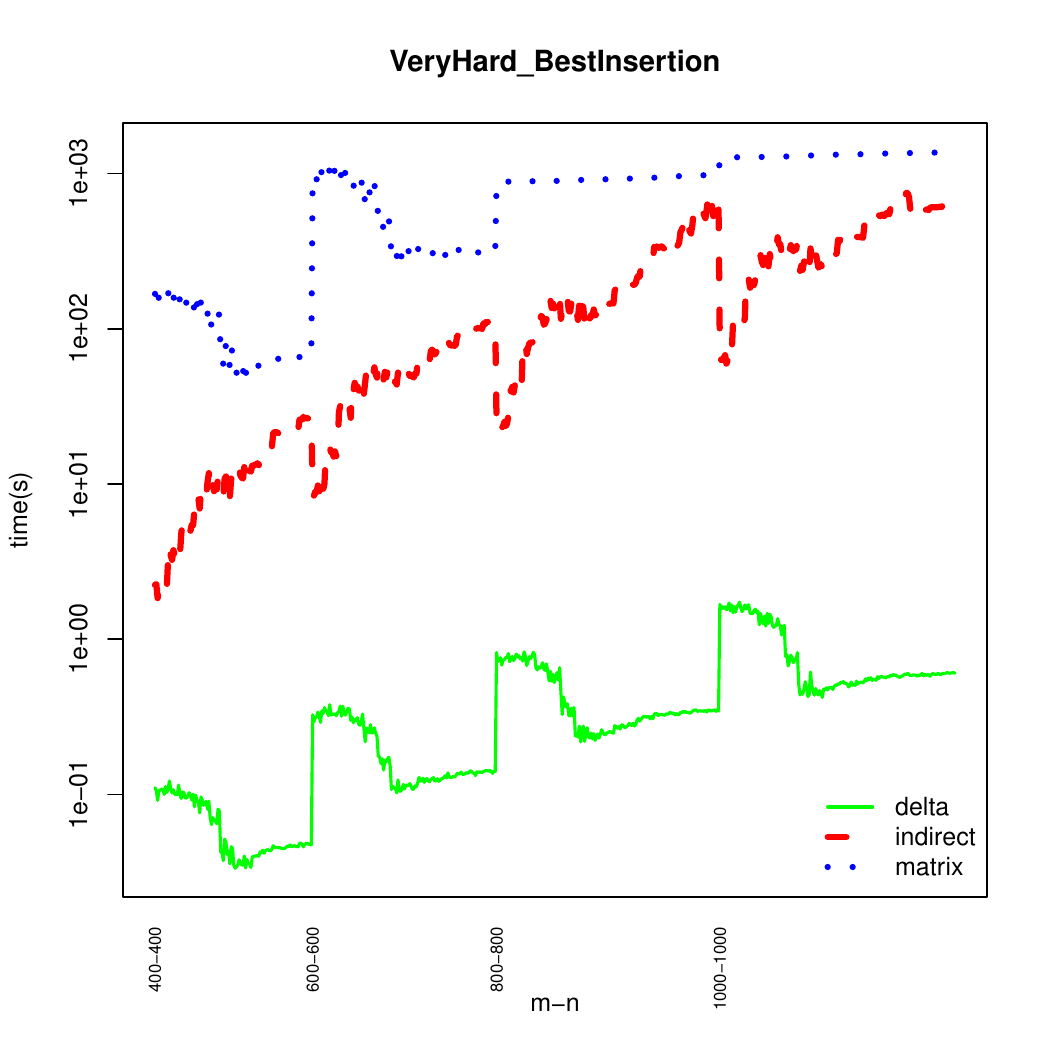}
  \caption*{(d)}
\endminipage
\caption{Time comparison of evaluation methods applied on a best insertion procedure}
\label{graphs_bestinsertion}
\end{figure}

This local search randomly selects one column per time and inserts it in every possible position of the solution. The column is fixed in the position that yields the best result value. Figure \ref{graphs_bestinsertion} shows the time comparison for the three evaluation methods. 

The $\Delta$-evaluation presents its best performance for this local search procedure. The dominance of the proposed evaluation method can be observed even for small instances on the Becceneri and Chu\&Stuckey instance set. The difference is even more evident in larger instances. For example, on the Very Hard instance set, the evaluation time starts at 0.11 seconds for the first 400$\times$400 instance and has its last measure of 0.61 seconds for the last 1000$\times$1000 instance; on the other hand, the indirect evaluation starts at 2.24 seconds and ends at 651.93 seconds for the same instances. The $\Delta$-evaluation also presents a lower time average in all the instance sets. The matrix evaluation, as expected, shows the worst performance in all the tests. 

The $\Delta$-evaluation method is also invoked less for the best insertion procedure. On average, the $\Delta$-evaluation was called 2,823 times, versus 1,121,407 times for the indirect and matrix evaluation. The Wilcoxon signed-rank test (V = 0, p-value $< 2.2 \cdot 10^{-16}$) indicates that there are statistical differences between the $\Delta$-evaluation and indirect evaluation method regarding the time performance on the Very Hard instance set. The value of $V$ equals zero indicates that all the $\Delta$-evaluation values are lower than the indirect evaluation values. Lastly, the $\Delta$-evaluation can be easily employed in the best insertion heuristic to refine and find local optimum, with a minimum increase in processing time for solutions generated by metaheuristics. This method can also evaluate insertions in fixed positions, maintaining the same time performance.

\subsection{2-Swap}
\begin{figure}[!ht]
\centering 
\minipage{0.45\textwidth}
  \includegraphics[width=\linewidth]{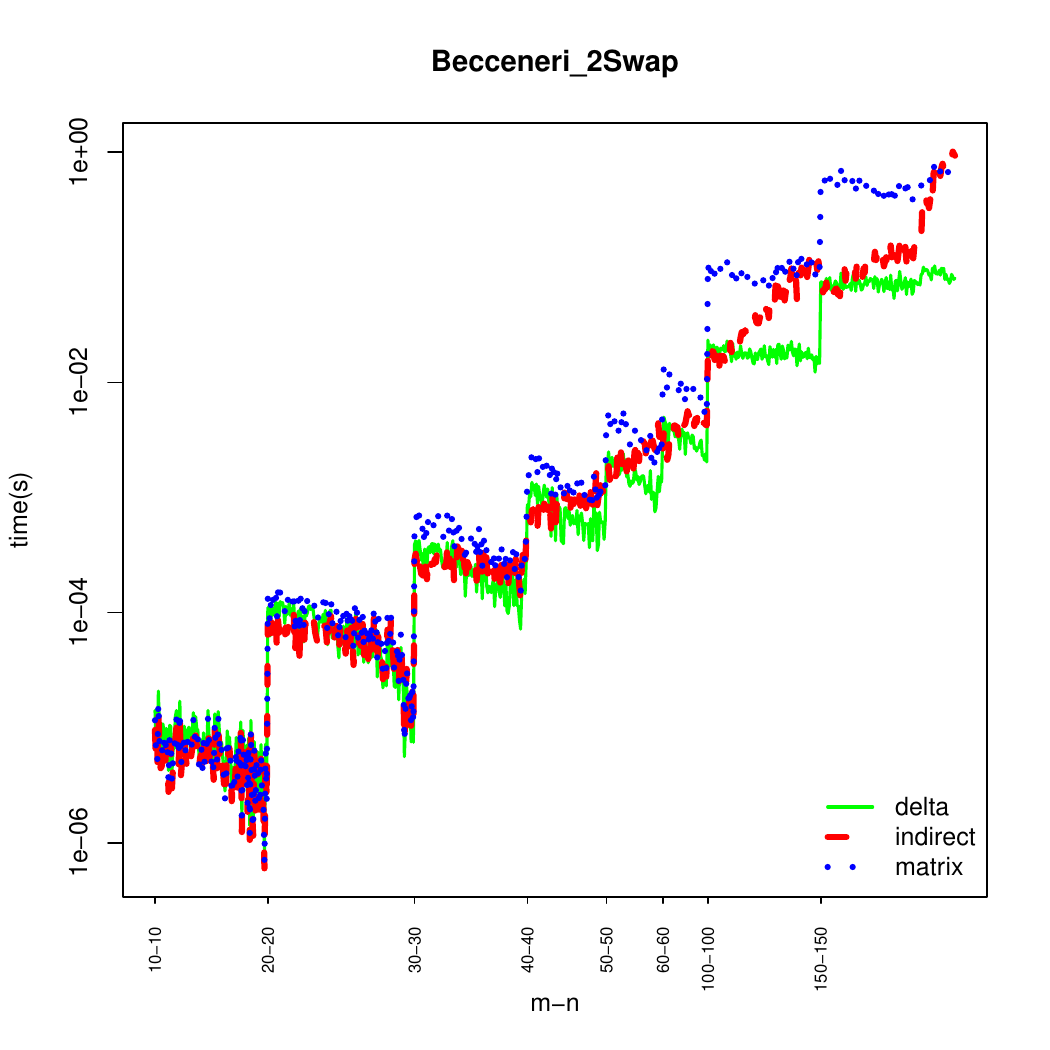}
  \caption*{(a)}
\endminipage
\minipage{0.45\textwidth}
  \includegraphics[width=\linewidth]{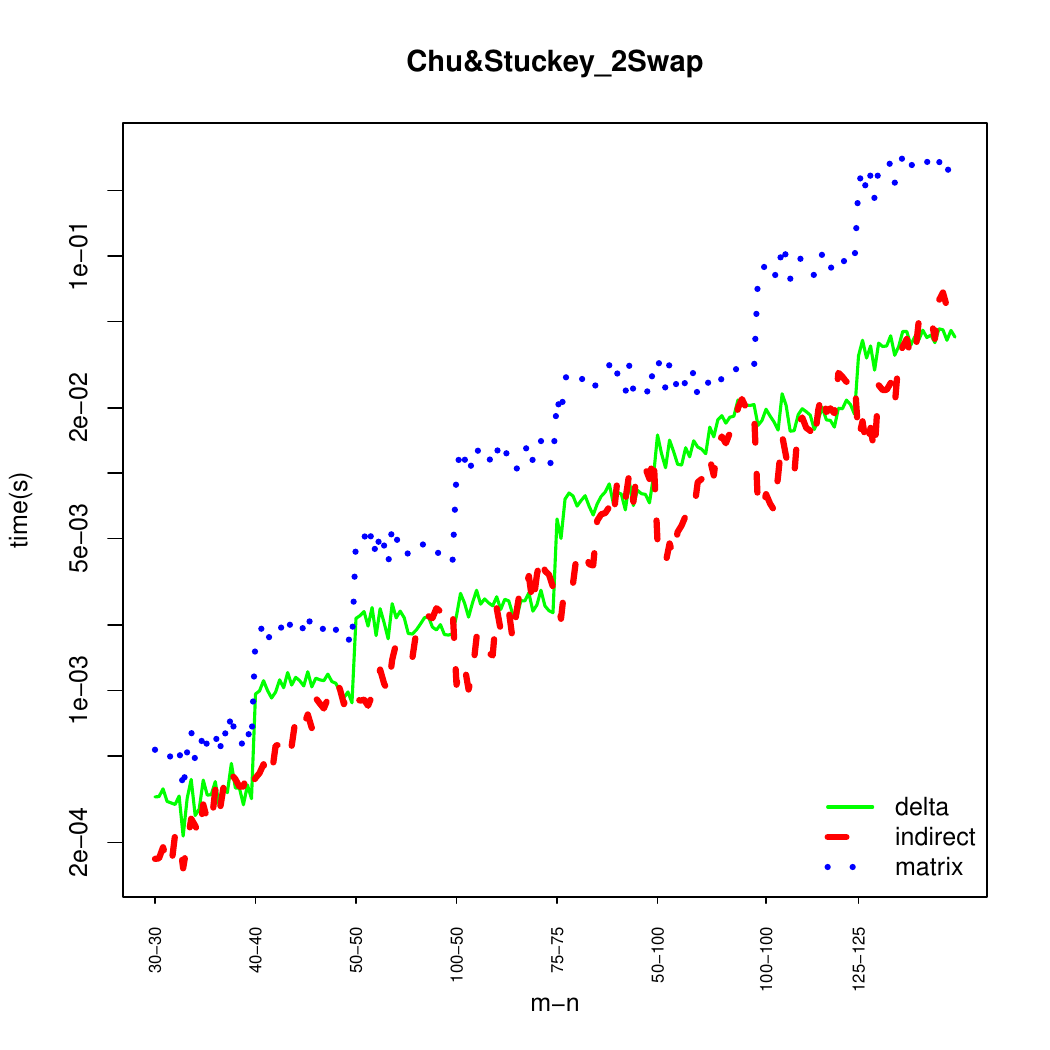}
  \caption*{(b)}
\endminipage
\\
\minipage{0.45\textwidth}
  \includegraphics[width=\linewidth]{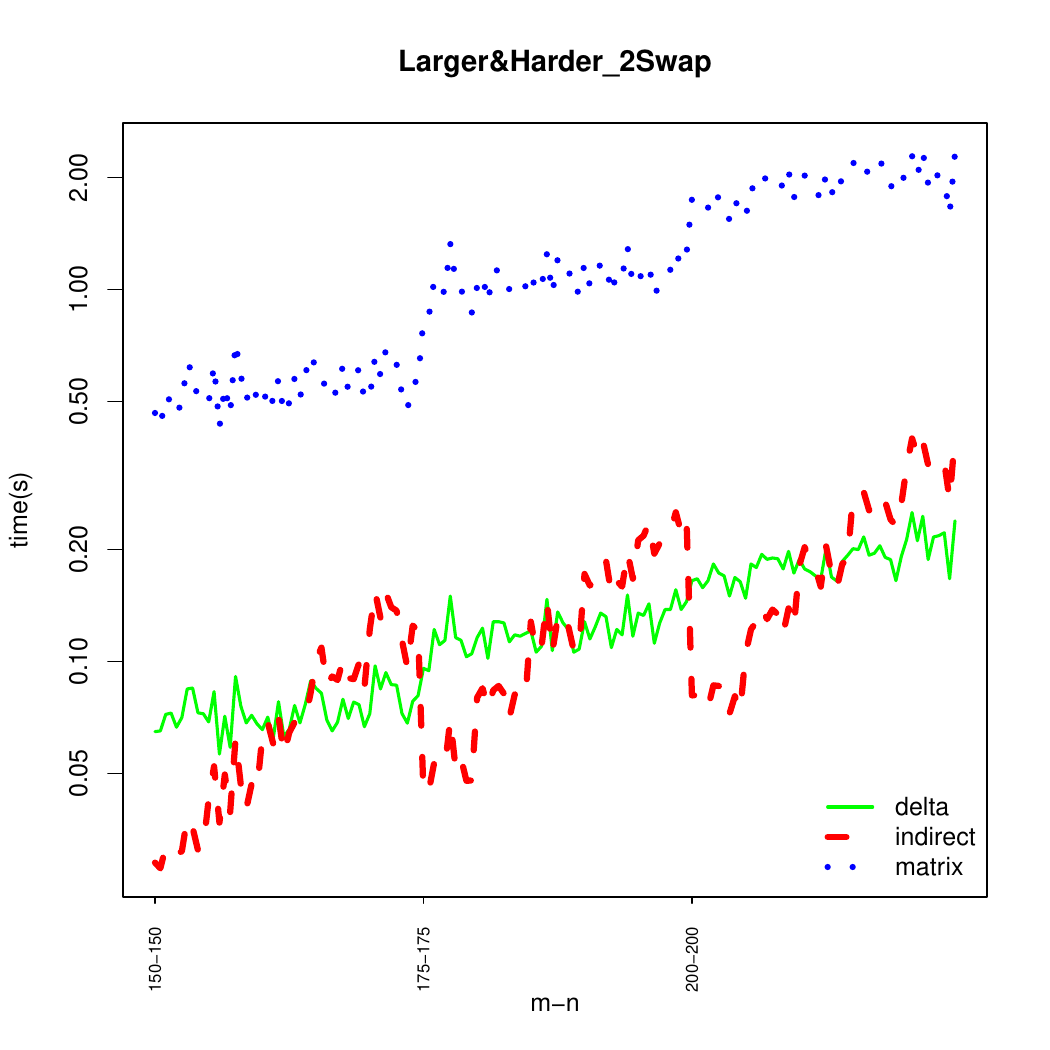}
  \caption*{(c)}
\endminipage
\minipage{0.45\textwidth}
  \includegraphics[width=\linewidth]{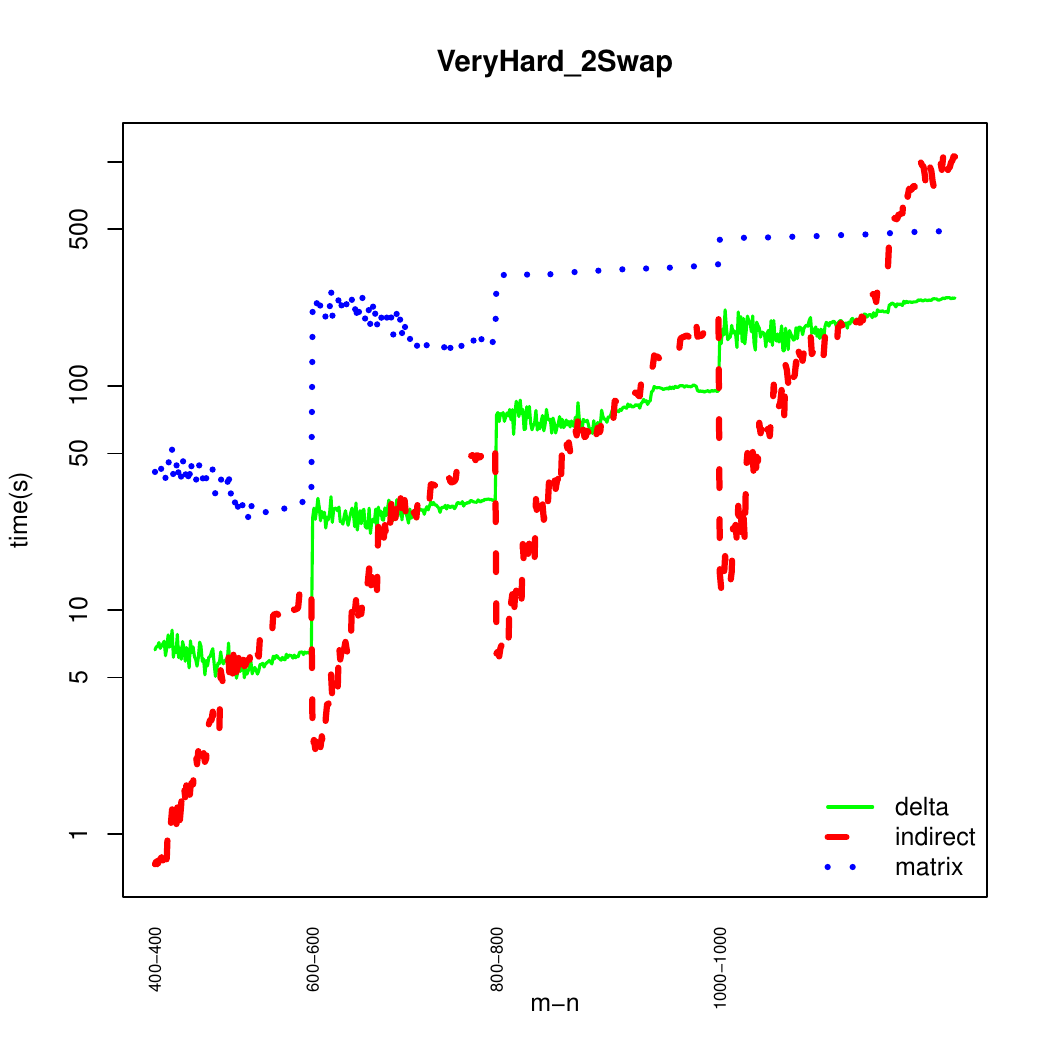}
  \caption*{(d)}
\endminipage
\caption{Time comparison of evaluation methods applied on a 2-swap procedure}
\label{graphs_2swap}
\end{figure}

All pairs of columns are generated and swapped for this local search in a random sequence. Figure \ref{graphs_2swap} shows the time comparison for the three evaluation methods. 

The $\Delta$-evaluation method makes two insertion movements for this local search. Hence, it requires more processing time. However, the method is still competitive, mainly for large and dense instances. The proposed method only has a superior time average for the Chu\&Stuckey instance set, outperforming the indirect and matrix evaluation method on the other three instance sets. 

The $\Delta$-evaluation can maintain a more linear processing time for instances of the same same, independent of its density. On the other hand, the indirect evaluation is faster for less dense instances and significantly increases time for more dense instances. For example, the indirect evaluation takes 14.52 seconds to process the first 1000$\times$1000 instance and 1053.36 for the last and more dense 1000$\times$1000 instance. The $\Delta$-evaluation takes 174.55 and 246.85 seconds for the same two instances. The matrix evaluation is not influenced as much by the instance density. However, its running time is, without question, much slower. 

The number of calls is the same for the three methods, given that the methods need to be invoked only once for each swap. The Wilcoxon signed-rank test (V = 96898, p-value $= 0.8036$) indicates no statistical difference between the $\Delta$-evaluation and indirect evaluation method regarding the time performance on the Very Hard instance set. However, the proposed method has a lower time average, is more consistent, and can be employed instead of the indirect method without performance loss.

\subsection{2-Opt}
\begin{figure}[!ht]
\centering 
\minipage{0.45\textwidth}
  \includegraphics[width=\linewidth]{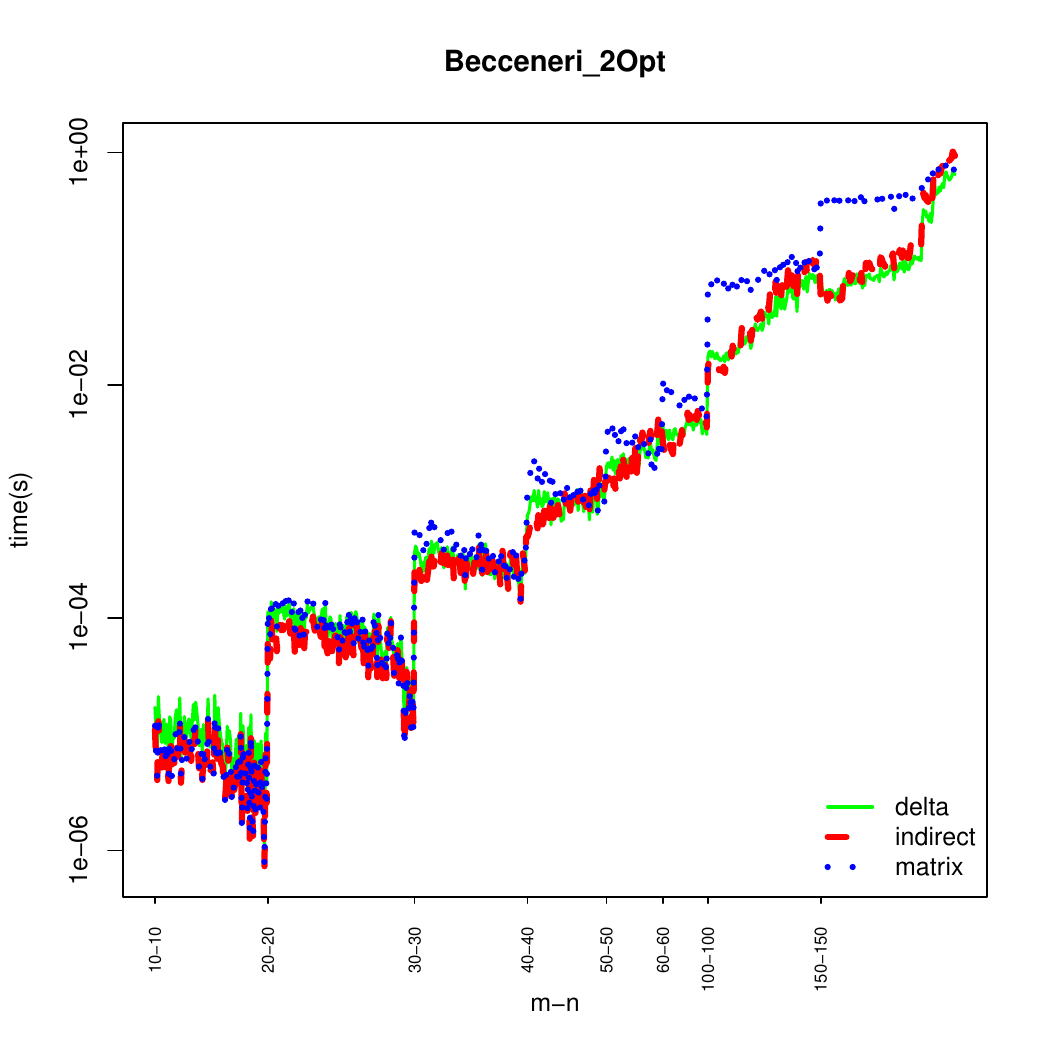}
  \caption*{(a)}
\endminipage
\minipage{0.45\textwidth}
  \includegraphics[width=\linewidth]{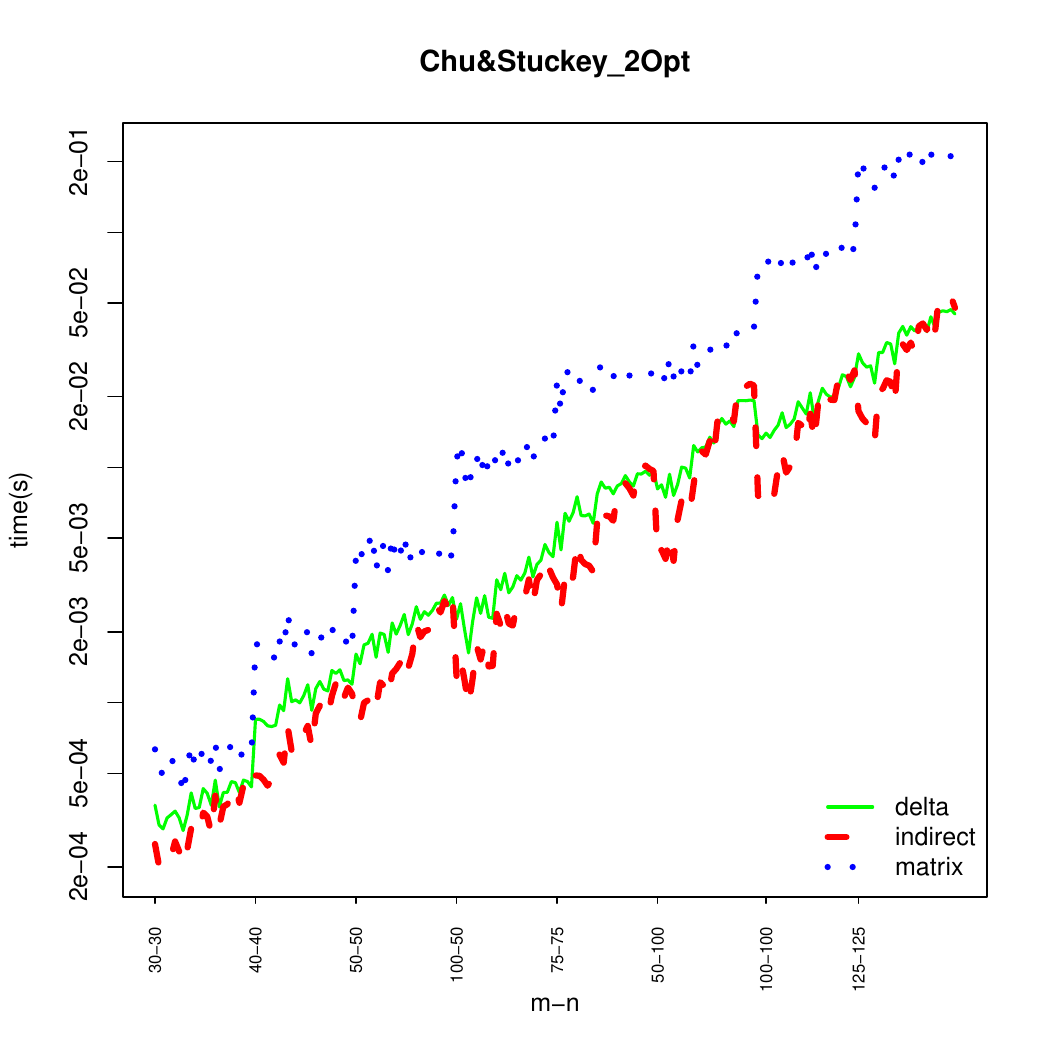}
  \caption*{(b)}
\endminipage
\\
\minipage{0.45\textwidth}
  \includegraphics[width=\linewidth]{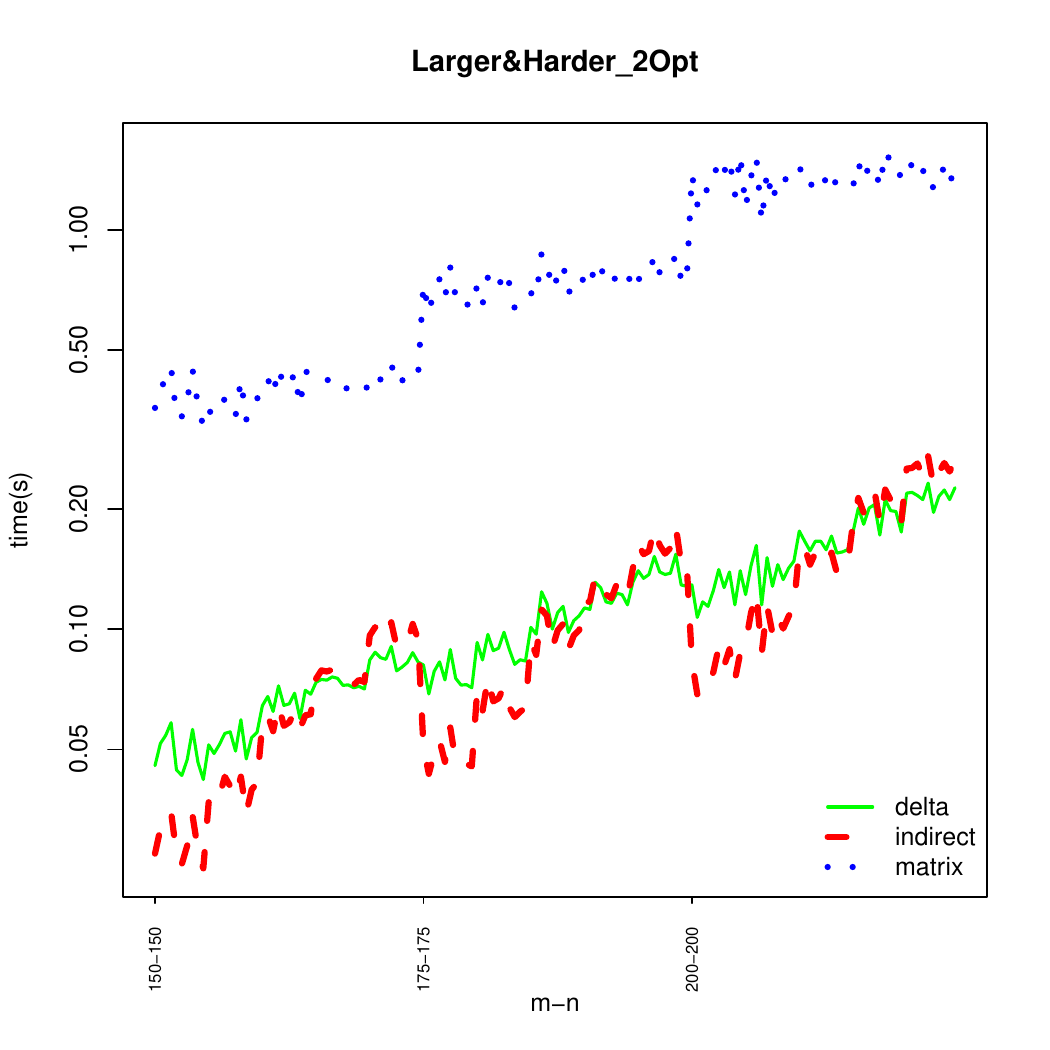}
  \caption*{(c)}
\endminipage
\minipage{0.45\textwidth}
  \includegraphics[width=\linewidth]{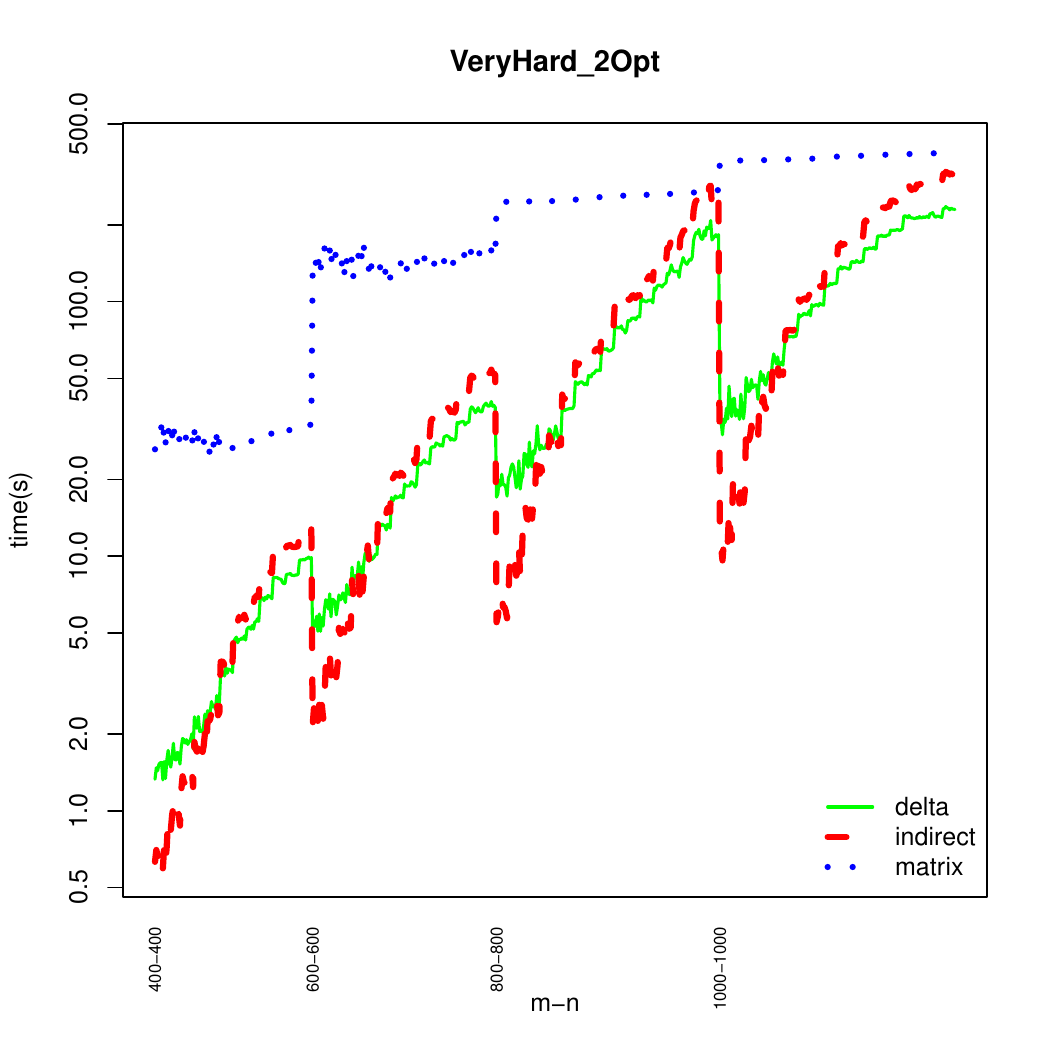}
  \caption*{(d)}
\endminipage
\caption{Time comparison of evaluation methods applied on a 2-opt procedure}
\label{graphs_2opt}
\end{figure}

The last local search also generates all pairs of columns. The procedure inverts the position of all the column sequences from the first column in the pair to the second one. Figure \ref{graphs_2opt} shows the time comparison for the three evaluation methods. 

\FloatBarrier
The $\Delta-$ evaluation method for the 2-opt procedure reevaluates only the portion of the solution that is being inverted. Therefore, it is slightly faster than the indirect evaluation method. The average time for each instance set is 0.010, 0.037, 0.11 and 65.30 for the $\Delta-$evaluation and 0.009, 0.049, 0.11, 79.44 for the indirect evaluation method. 

Similarly to the performance on the 2-swap procedure, the $\Delta-$ evaluation presents its best performance in large and more dense instances for the 2-opt local search. The number of calls is the same for the three methods. The matrix evaluation is once more the slower of the three methods and is not recommended for use.

The Wilcoxon signed-rank test (V = 39310, p-value $< 2.2 \cdot 10^{-16}$) indicates that there are statistical differences between the $\Delta$-evaluation and indirect evaluation method regarding the time performance on the Very Hard instance set.

\section{Conclusion}
\label{conclusions}

We have addressed a generic column permutation problem on binary matrices that produce permutation matrices that hold the consecutive ones property. This generic problem models many theoretical and real-world optimization $\mathcal{NP}$-Hard problems from different contexts such as graph theory, very large-scale integration design, and industrial production planning. A new evaluation method, $\Delta$-evaluation, was proposed and proved to be faster in two out of three test scenarios compared to more intuitive evaluation methods. The average running time is lower in almost all the instance sets used in the computational experiments. The proposed method is significantly faster for larger and more dense instances. Statistical tests proved that the proposed method outperforms the more intuitive methods in two commonly used local search procedures. The proposed method can be employed in descent local search procedures, especially for the best insertion method, with a minimum time cost. 

\clearpage

%----------------------------------------------------------------------------------------
%	Bibliography
%----------------------------------------------------------------------------------------

\end{document}